\documentclass[conference]{IEEEtran}
\IEEEoverridecommandlockouts
\usepackage{cite}
\usepackage{amsmath,amssymb,amsfonts}
\usepackage{algorithmic}
\usepackage{graphicx}
\usepackage{textcomp}
\usepackage{xcolor}
\def\BibTeX{{\rm B\kern-.05em{\sc i\kern-.025em b}\kern-.08em
    T\kern-.1667em\lower.7ex\hbox{E}\kern-.125emX}}
\begin{document}

\title{Indoor Positioning using Wi-Fi and Machine Learning for Industry 5.0\\

}

\author{\IEEEauthorblockN{ Inoj Neupane}
\IEEEauthorblockA{\textit{School of Computer (CDMS)} \\
\textit{Western Sydney University}\\
Sydney, Australia \\
}
\and
\IEEEauthorblockN{ Belal Alsinglawi}
\IEEEauthorblockA{\textit{School of  Computer (CDMS)} \\
\textit{Western Sydney University}\\
Sydney, Australia \\
}
\and
\IEEEauthorblockN{ Khaled Rabie}
\IEEEauthorblockA{\textit{Department of Engineering
} \\
\textit{Manchester Metropolitan University}\\
Manchester, UK \\
}
}

\maketitle

\begin{abstract}
Humans and robots working together in an environment to enhance human performance is the aim of Industry 5.0. Although significant progress in outdoor positioning has been seen, indoor positioning remains a challenge. In this paper, we introduce a new research concept by exploiting the potential of indoor positioning for Industry 5.0. We use Wi-Fi Received Signal Strength Indicator (RSSI) with trilateration using cheap and easily available ESP32 Arduino boards for positioning as well as sending effective route signals to a human and a robot working in a simulated-indoor factory environment in real-time. We utilized machine learning models to detect safe closeness between two co-workers (a human subject and a robot). Experimental data and analysis show an average deviation of less than 1m from the actual distance while the targets are mobile or stationary.
\end{abstract}


\section{Introduction}
Industry 5.0 aims to enhance human activities through collaboration between humans and robots in various applications such as manufacturing and supply chain management \cite{b1}. Location-based services play an important role in these applications. In outdoor scenarios, Global Positioning System (GPS) is used for effective positioning, but it is not applicable in indoor environments due to obstruction of indoor objects and absorption from walls that attenuate GPS signals. There are several wireless technologies available for indoor positioning, such as Bluetooth, Radio-Frequency Identification (RFID), Ultra-wideband (UWB), Zigbee, Sigfox, and LoRa, but Wi-Fi is gaining popularity due to its widespread availability and no added cost. Among several methods available for indoor positioning, Wi-Fi accounts for almost 50\% of studies published on this topic \cite{b2}.

Lateration is a positioning method that involves measuring distances from reference nodes to determine the location of an unknown object or anchor point. At least three reference nodes are needed for trilateration. When more than three reference nodes are used, it is called multilateration \cite{b3}. RSSI is more commonly used than Channel State Information (CSI) for Wi-Fi positioning because it requires less specific wireless network interface cards \cite{b4}. Wi-Fi RSSI is a widely available and cost-effective solution. Additionally, trilateration is one of the simplest and most economical methods to implement for indoor positioning \cite{b5}.

Wi-Fi RSSI fingerprinting technique using Recurrent Neural Networks (RNN) was studied in \cite{b4}. Although the results showed significant accuracy with an average error of 0.75m and 80\% errors under 1m, the data collection process for fingerprinting and training neural networks is costly and labor-intensive. In \cite{b5}, RFID tags were used for indoor positioning, which required the use of RFID beacons placed in specific positions. The authors in \cite{b6} used a pair of radio-frequency transceivers specifically designed for precise distance measurement. Bluetooth Low Energy (BLE), a new technology with extremely low power consumption, was used effectively in \cite{b7}, but its implementation for indoor positioning requires additional resources for acting as reference nodes that are generally not required in indoor environments for other purposes. UWB technology was used in \cite{b8}, achieving centimeter-level accuracy, but it requires expensive hardware equipment.

\begin{figure*}
\centerline{\includegraphics[width=7.31in]{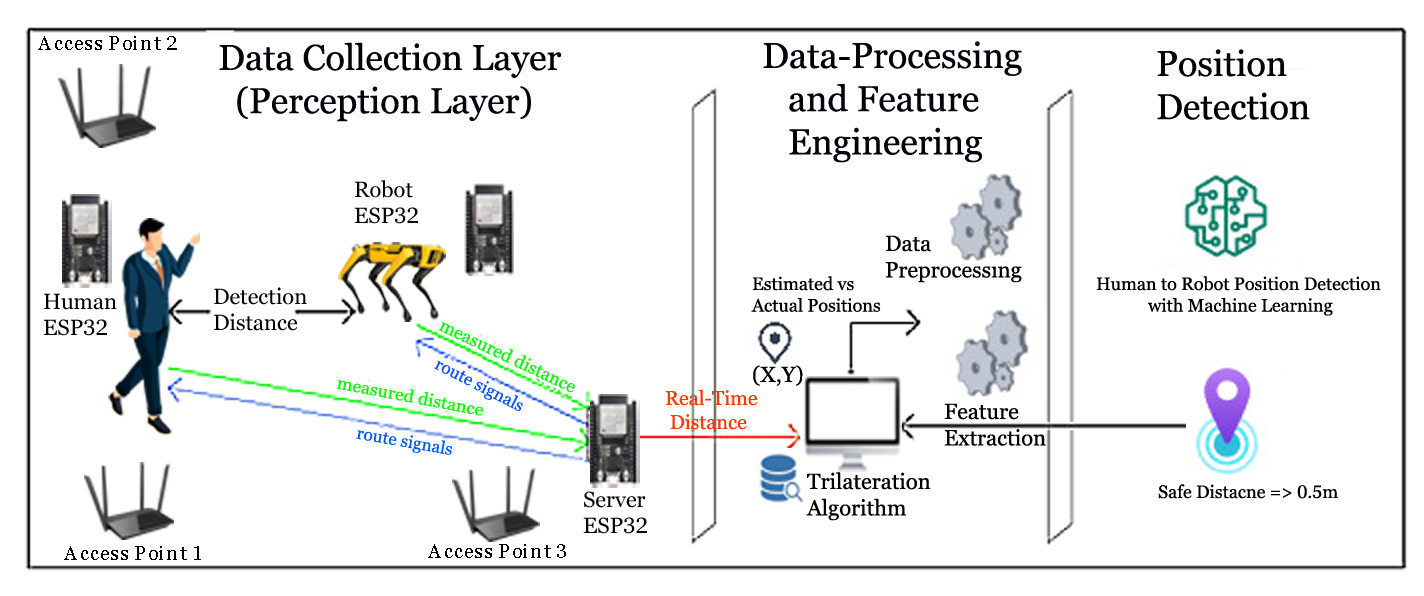}}
\caption{The IPS-Industry 5.0 system architecture.}
\label{fig1}
\end{figure*}

In contrast to previous studies, we performed positioning in a simulated smart factory environment using Wi-Fi RSSI trilateration and ML to further process the data. We used ESP32 boards with built-in WiFi capability. One board was attached to a human and another to a robot as anchor points in the setup. The boards received RSSI values from three readily available WiFi access points and sent distance values to a central third ESP32 board server. The central ESP32 server performed trilateration and provided the location coordinates of the human subject and robot. Whenever they were within a certain proximity threshold, simulating a potential collision, a signal was produced. Additionally, we implemented ML to further strengthen the signal for navigation purposes as RSSI values are highly susceptible to noise.

The main contributions of this work are as follows:
\begin{enumerate}
\item This work introduces a complete study for indoor localization using readily available, ESP 32 boards and WiFi Access Points to localize a human subject and robot in Industry 5.0 setting.
\item This study presents the implementation of ML methods to further process the data to make reliable navigation decisions.
\item One of the challenges of Wi-Fi indoor localization, specifically, the short collection time of RSSI per location, has been addressed and discussed.
\end{enumerate} 

The paper is organized as follows. In Section 2, the theoretical framework is presented, discussing the role of each layer. Section 3 describes the practical implementation, including hardware, software, and experimental procedure description. Section 4 presents the results obtained in different scenarios, and section 5 concludes the paper with future research directions.

\section{The IPS-Industry 5.0 System}
Figure 1 illustrates the IPS-Industry 5.0 system architecture. The IPS-Industry 5.0 operates in three phases: a data collection layer, a data pre-processing and cleaning layer, and a position detection layer, as described in the following subsections.

\subsection{Data Collection Layer}
Lateration is a method used for calculating the coordinates of a point by using distances from a series of known points. When three references or known points are used, the method is referred to as trilateration. A simplified formula for relating the Wi-Fi RSSI with the distance between the reference nodes and the anchor points \cite{b7}is:

\begin{equation}
\emph{P} = \emph{A} - 10\emph{n}\normalfont{log}(\emph{d})\label{eq1}
\end{equation}
where, \emph{P} represents the RSSI in dBm, \emph{A} is the RSSI (dBm) value received from an access point when the distance of the receiver is 1m, \emph{n} is the environmental factor and \emph{d} is the distance between the transmitter or reference nodes, and receiver or anchor points. 

The first step is to measure the value of \emph{A} parameter by placing the receiver at 1m from the transmitter and measuring the RSSI values. The second step is to measure the value of \emph{n} between the transmitter and receiver. For this, (1) is modified to determine the value of \emph{n}. The receiver is placed at a known distance and the measured \emph{A} parameter is used. Finally, after determining the values of \emph{A} and \emph{n}, the distance can be calculated using (1).

For the positioning, as in \cite{b7} the access points are precisely located at locations (0, 0), (0, \emph{y2}), and (\emph{x3}, 0) to provide a simplified coordinate as:
\emph{
\begin{equation}
{(x, y)}  = \left(\frac{x_1^2 + (d_1^2 - d_2^2)}{2x_2}, \frac{y_2^2 + (d_1^2 - d_3^2)}{2y_2}\right)\label{2}
\end{equation}
}
where \emph{$d_1$, $d_2$}, and \emph{$d_3$} represents the distance from access points 1, 2, and 3 respectively.

This layer generates the distance between the human and each of the three access points (H-D1, H-D2, and H-D3), as well as the distance between the robot and each of the three access points (R-D1, R-D2, and R-D3), measured in meters. These distances are then used as input for the trilateration algorithm, which is used to calculate the real-time position of both the human (Hx, Hy) and the robot (Rx, Ry) in relation to the access points. The algorithm also calculates the estimated distance between the human and the robot in both stationary and real-time movements. The stationary data collection was recorded over a 90-second period, while a 5-second interval was used for three scenarios of real-time positioning and detecting proximity between the human and robot.

\subsection{Data Pre-processing and Features Engineering}
We use ML to extract new features that indicate the error in estimated positions relative to the actual positions (ground truth) of the human and robot in both stationary and real-time scenarios. This allows us to determine if the robot and human's positions are close enough to trigger an alarm message (envisioned), as the trilateration method alone is susceptible to uncertainty due to environmental interference. This improves the performance of the system and enables more informed decisions on human-robot movements and presence in Industry 5.0.

\subsection{Detecting Human-Robot proximity with ML}
In this phase, the proximity between humans and robots was treated as a classification problem with two binary outcomes: 0 (not close) and 1 (close). A threshold of 0.5m was established, meaning that if the distance between the human and the robot was less than 0.5m, the robot was considered close enough to the human, and if it was greater than 0.5m, a safe distance existed between them in the context of Industry 5.0 intelligent spaces. To achieve threshold detection (0.5 m), three supervised machine learning models were employed. We utilized logistic regression \cite{b9} as a binary classification problem for predicting human-robot distance proximity, to predict the probability that an instance belongs to the default class (as 0 or 1). Additionally, we implemented Stochastic Gradient Descent (SGD) \cite{b10}, which supports different loss functions and penalties for classification problems. The decision boundary of an SGD classifier is trained with the hinge loss. Lastly, we used a Linear Support Vector Classifier (SVC) \cite{b11} with a linear kernel function to perform the classification prediction task. Performance was evaluated using accuracy, which is the percentage of correct predictions, and F1-Score, which is a measure of precision and recall. Precision is the number of true positive results divided by the total number of positive results, including those that were incorrectly identified, while recall is the number of true positive results divided by the total number of samples that should have been identified as positive.

\section{Implementation}

\subsection{Hardware}
The ESP32 is a single chip that combines Wi-Fi and Bluetooth technology, making it ideal for use in wearable and the Internet of Things (IoT) applications. Its ultra-low power consumption and complete integration solution make it a great option. Additionally, its compliance with Bluetooth Low Energy (BLE) allows for the easy transfer of data to other low-energy devices for further processing. 

In our experiment, we utilized three ESP32s: one was attached to a human subject, another was attached to a robot, and the third served as a central receiver. The central ESP32, which was attached to a computer, was used to analyze the real-time location and proximity details, while the other ESP32s were used to perform trilateration on the received distance values from the human and robot with respect to different access points.

\subsection{Software}
The Arduino IDE offers a free text editor for writing code, along with a variety of functions and menus to develop software for Arduino devices. In our experiment, we programmed two ESP32s, which were attached to a human and a robot, to take samples of Wi-Fi RSSI from three access points at intervals of 0.1 seconds. The third ESP32 was programmed to receive the distance values from the other two ESP32s and to calculate coordinates using trilateration, as well as the distance between the human and robot. If the proximity between the devices was measured to be beyond a certain threshold, the program updated a flag for navigation purposes.

\begin{figure}[t]
\centerline{\includegraphics[width=3.48in]{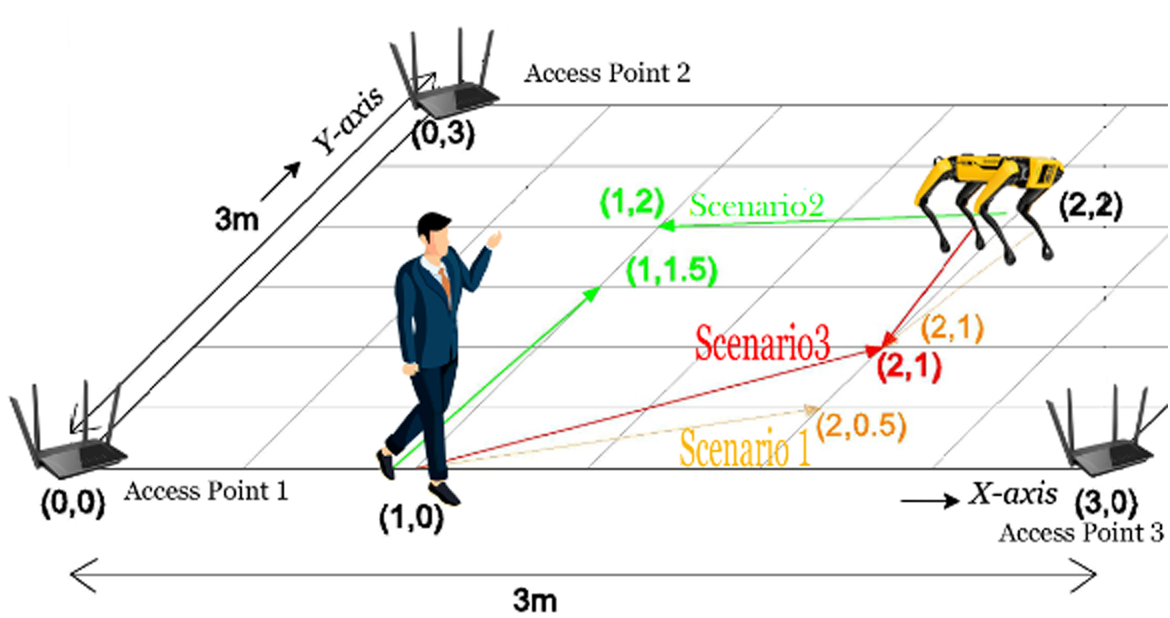}}
\caption{Different cases of experiments: stationary and mobile scenarios.}
\label{fig2}
\end{figure}

\subsection{Experimental Procedure}
The experiment was conducted at Western Sydney University in a 3m x 3m grid area. The environment simulated a real-life scenario with interfering objects and signals. First, for each access point, the \emph{A} and \emph{N} parameters were calculated by averaging more than 500 samples per access point for each parameter. The averaged parameters were then coded into the ESP32s attached to the human and robot to obtain the distance values from each access point. To perform real-time distance calculations, the speed of RSSI sampling was increased by configuring the ESP32's WIFI receiver to only tune to the channels of the used access points. A sample time of 100ms was chosen as it resulted in the least amount of RSSI value loss while still maintaining a clear signal.

The ESP32s attached to the human subject and robot calculated the distance and then sent the values to the central ESP32 server using the ESPNOW protocol, which is readily available in the Arduino IDE. The central ESP32 then performed trilateration on the received distances and displayed the location information. If a certain threshold proximity was detected between the human subject and robot, a signal was produced and sent to the human and robot for navigation purposes.

\section{Results}
The first set of data was collected while the human subject and robot targets were stationary at locations (1,0) and (2,2) respectively, as shown in Figure 2. More than 900 samples were analyzed, and the results showed an accuracy of 70.99\% with an average deviation of the distance between the human and robot of 0.65m. The second set of data was collected while the human subject moved towards the position (2,0.5) and the robot moved towards the position (2,1), as represented by scenario 1 (orange color) in Figure 2. The results showed an accuracy of 28.3\% with an average deviation of the distance between the human subject and robot of 0.89m. The third set of data was collected while the human subject moved towards the position (1,1.5) and the robot moved towards the position (1,2), as represented by scenario 2 (green color) in Figure 2. The results showed an accuracy of 42.85\% with an average deviation of the distance between the human subject and robot of 0.75m. The final set of data was collected while both the human and robot moved towards the position (2,1) simulating a collision scenario as represented by scenario 3 (red color) in Figure 2. The results showed an accuracy of 36.91\% with an average deviation of the distance between the human subject and robot of 0.77m.

In the ML detection stage, models with an F1-Score of at least 0.5 for recognizing the safe distance between a human and a robot demonstrated relatively high accuracy. Linear-SVC outperformed other models in terms of accuracy and F1-Score for stationary positioning, achieving 73\% and 74\% respectively. In scenario 1, the accuracy and F1-score of LR and SGD were 91\% and 93\% respectively. In scenario 2, all models performed optimally with perfect precision and F1-Score. Scenario 3 demonstrated that LR performed better than other models in terms of accuracy and F1-Score (75\% and 78\% respectively). These results indicate that LR is a superior classifier for assessing the proximity of human-to-robot communication in industrial 5.0 spaces, particularly for fixed and real-time indoor locations. Notably, our identification process for a safe distance between the human subject and a robot was carried out using offline-collected data. 

\begin{table}[htbp]
\caption{Results for Each Case for ML}
\begin{center}
\begin{tabular}{ ||c|c|c|  }
\hline
MODEL& ACCURACY \%& F1-SCORE\\
\hline
\multicolumn{3}{|l|}{Stationary Case:} \\
\hline
LR& 45& 0.63\\
\hline
SGD& 55& 0.69\\
\hline
Linear-SVC& 73& 0.81\\
\hline
\multicolumn{3}{|l|}{Scenario 1:} \\
\hline
LR& 91& 0.93\\
\hline
SGD& 91& 0.93\\
\hline
Linear-SVC& 87& 0.90\\
\hline
\multicolumn{3}{|l|}{Scenario 2:} \\
\hline
LR& 100& 1\\
\hline
SGD& 100& 1\\
\hline
Linear-SVC& 100& 1\\
\hline
\multicolumn{3}{|l|}{Scenario 3:} \\
\hline
LR& 75& 0.78\\
\hline
SGD& 44& 0.46\\
\hline
Linear-SVC& 50& 0.53\\
\hline
\end{tabular}
\label{tab1}
\end{center}
\end{table}

\section{Conclusion}
A novel, accurate indoor positioning system for Industry 5.0 was presented. The preliminary tests of indoor positioning with navigation purposes between a human and a robot in the context of Industry 5.0 were conducted using easily available setups that employed trilateration and machine learning (ML) techniques. Fluctuations were observed when both the human and robot were moving due to interference from WiFi signals with human and robot bodies. However, the ML showed significant results in simulated situations of closer proximity, mitigating those fluctuations. The average deviation of the distance between the human and the robot was found to be less than 1m in all scenarios.  

In the near future, our current work will focus on developing the online detection of a safe distance. We plan to extend the duration of data collection and incorporate more complex scenarios in which robots must make decisions to avoid collisions, particularly in the dead zone regions of Industry 5.0 intelligent space. It is highlighted that the models' F1 score for scenario 2 was optimum. As a result, the data does not adequately represent real-world problems, making it challenging for models to discover useful patterns. We expect to collect more data to increase its representativeness. In addition, we intend to employ regularization to prevent overfitting in the extended version of this study with more collected data points, as well as analyze the model's performance during the online prediction phase. We will use increasingly advanced models such as deep neural networks to account for the intricacy of the spatio-temporal relationships between the robot and the human. Furthermore, we will investigate the application of the Kalman filter and the fusion of other systems, such as IMU, to increase the overall effectiveness of the system. Finally, we intend to capitalize on the advancements in WiFi positioning that are based on time of flight, as inspired by recent research in the field \cite{b12}.

\end{document}